\documentclass[sigconf]{acmart}

\AtBeginDocument{%
  \providecommand\BibTeX{{%
    \normalfont B\kern-0.5em{\scshape i\kern-0.25em b}\kern-0.8em\TeX}}}

\copyrightyear{2020}
\acmYear{2020}
\setcopyright{acmcopyright}
\acmConference[MM '20]{Proceedings of the 28th ACM
	International Conference on Multimedia}{October 12--16, 2020}{Seattle, WA, USA}
\acmBooktitle{Proceedings of the 28th ACM International Conference on Multimedia
	(MM '20), October 12--16, 2020, Seattle, WA, USA}
\acmPrice{15.00}
\acmDOI{10.1145/3394171.3413900}
\acmISBN{978-1-4503-7988-5/20/10}

\usepackage{multirow}
\usepackage{makecell}
\usepackage{diagbox}
\usepackage{footnote}
\usepackage{times}
\usepackage{balance}

\begin{document}
\fancyhead{}

\title{TRIE: End-to-End Text Reading and Information Extraction for Document Understanding}


\author{Peng Zhang$^{1*}$,Yunlu Xu$^{1*}$,Zhanzhan Cheng$^{21\dagger}$,Shiliang Pu$^{1+}$,Jing Lu$^1$,Liang Qiao$^1$,Yi Niu$^{1}$, Fei Wu$^2$}
\thanks{\textsuperscript{*}Both authors contributed equally to this research. }
\thanks{\textsuperscript{$\dagger$}This work is completed under the supervision of Zhanzhan Cheng (contact email: 11821104@zju.edu.cn)}
\thanks{\textsuperscript{+}Corresponding author.}
\affiliation{
  \textsuperscript{1}\institution{Hikvision Research Institute, Hangzhou, China}
}
\email{(zhangpeng23,xuyunlu,pushiliang.hri,lujing6,qiaoliang6,niuyi)@hikvision.com}
\affiliation{
  \textsuperscript{2}\institution{Zhejiang University, Hangzhou, China}
}
\email{11821104@zju.edu.cn,wufei@cs.zju.edu.cn}


\begin{abstract}
Since real-world ubiquitous documents (\textit{e.g.,} invoices, tickets, resumes and leaflets) contain rich information, automatic document image understanding has become a hot topic. 
Most existing works decouple the problem into two separate tasks, (1) \textit{text reading} for detecting and recognizing texts in images and (2) \textit{information extraction} for analyzing and extracting key elements from previously extracted plain text.
However, they mainly focus on improving \textit{information extraction} task, while neglecting the fact that \textit{text reading} and \textit{information extraction} are mutually correlated.
In this paper, we propose a unified \textit{end-to-end text reading and information extraction network}, where the two tasks can reinforce each other. 
Specifically, the multimodal visual and textual features of \textit{text reading} are fused for \textit{information extraction} and in turn, the semantics in \textit{information extraction} contribute to the optimization of \textit{text reading}.
On three real-world datasets with diverse document images (from fixed layout to variable layout, from structured text to semi-structured text), 
our proposed method significantly outperforms the state-of-the-art methods in both efficiency and accuracy.
\end{abstract}

\ccsdesc[500]{Applied computing~Document analysis}
\ccsdesc[500]{Applied computing~Optical character recognition}
\ccsdesc[500]{Computing methodologies~Information extraction}

\keywords{End-to-End; Text Reading; Information Extraction; Visually Rich Documents}

\maketitle

\begin{savenotes}
\begin{table}[h]
	\caption{Example of VRDs and their categories.}
	\label{table:dataset_summary}
	\begin{tabular}{c|l|l}
		\diagbox{Layout}{\makecell[c]{Text\\type}} & \multicolumn{1}{c|}{Structured} & \multicolumn{1}{c}{Semi-structured} \\ \hline
		Fixed    & \makecell[l]{invoice, passport,\\ID card} & \makecell[l]{business email, \\sales contract}         \\ \hline
		Variable & \makecell[l]{purchase receipt, \\business card}  & \makecell[l]{resume, \\financial report}
	\end{tabular}
\end{table}
\end{savenotes}
\begin{figure}[h]
	\includegraphics[width=0.45\textwidth]{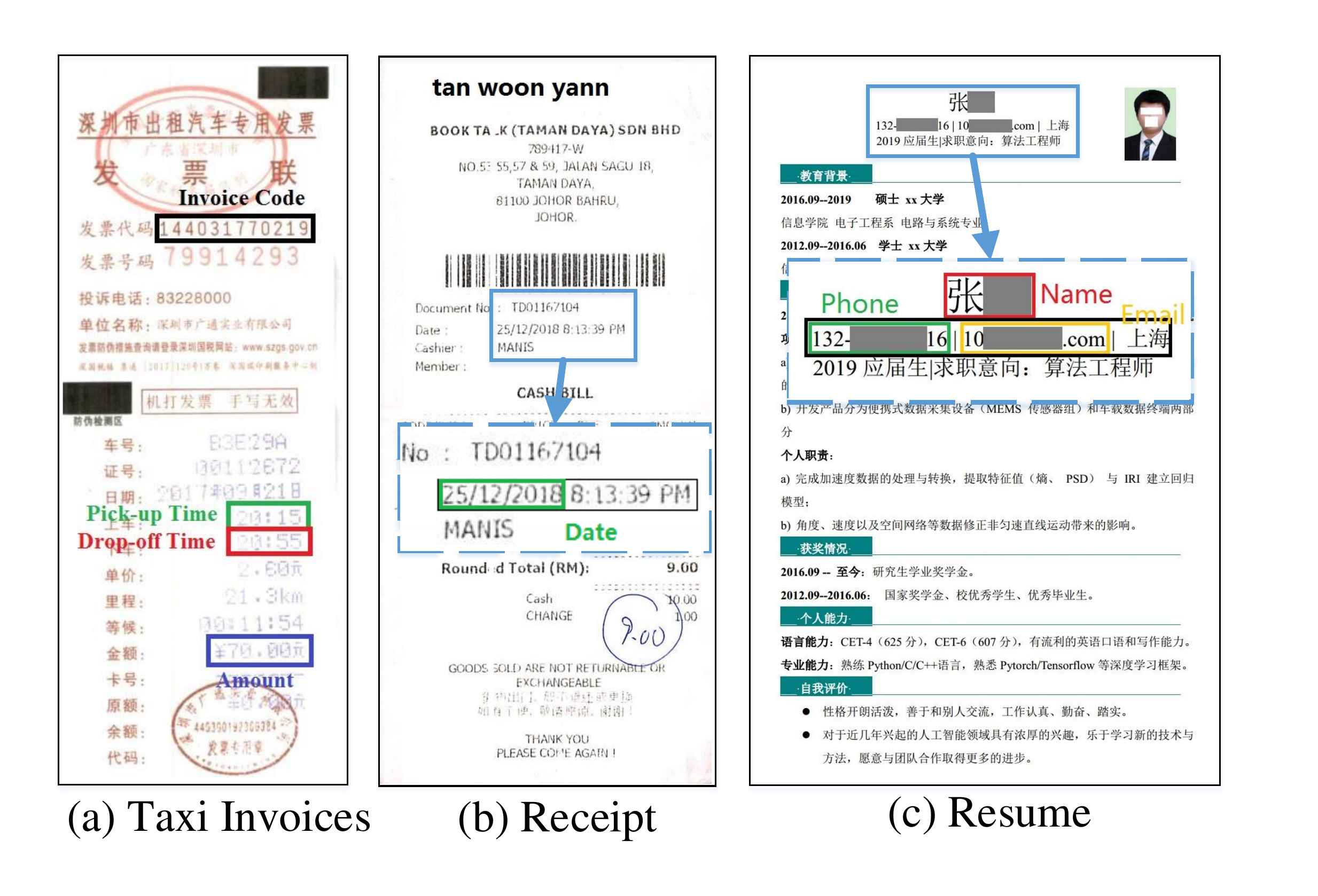}
	\caption{Examples of VRDs and example entities to extract. Sensitive information like name and contact information in resumes are pixelated to preserve privacy. Best viewed in color.}
	\label{fig:dataset}
\end{figure}
\section{Introduction}
Document understanding is a relatively traditional research topic that refers to the techniques to automatically handle with the text content. 
Among different types of documents, Visually Rich Documents (VRDs) have attracted more and more attention.
It is named after its composed modality of both text and vision that offers abundant information, including but not limited to layout, tabular structure, font size and even the font color in addition to plain text. 
Here, we divide them into \textit{four categories} from the dimensions of layout and text type.
\textit{Layout} here is defined as the relative positions of texts and \textit{text type} (\textit{i.e.,} structured, and semi-structured) follows the conventions of \cite{judd2004apparatus,soderland1999learning}.
In detail, \textit{Structured} data connotes data that is organized in a predetermined schema (\textit{e.g.,} invoices and receipt), while \textit{semi-structured} data denotes data that has one or more identifiers, but each portion of data is not necessarily organized in predefined fields, \textit{e.g.,} resumes.
Table~\ref{table:dataset_summary} summarizes common examples of VRDs and some examples are shown in Fig.~\ref{fig:dataset}. 
Automaticly recognizing texts and extracting valuable contents from VRDs can faciliate information entry, retrieval and compliance check. It is of great benefit to office automation in areas like accounting, financial and much broader real-world applications.

The complex problem of understanding VRDs is always decoupled into two different stages,
text reading and information extraction.  
Specifically, \textbf{text reading} includes text detection and recognition in images, which belongs to the optical character recogtion (OCR) research field and has already been widely used in many Computer Vision (CV) applications \cite{wang2020all,qiao2020textperceptron,feng2019textdragon}.
In the \textbf{information extraction} (IE) stage, specific contents (\textit{e.g.} entity, relation) are mined and processed from previously recognized plain text for diverse Natural Language Processing (NLP) tasks, such as Named Entity Recognition (NER) \cite{nadeau2007survey,lample2016neural,ma2019end} and Question-Answer (QA) \cite{yang2016stacked,anderson2018bottom,fukui2016multimodal}. 
Note that in existing routines, the two stages are separately executed. It means that the former recognizes text from images without semantic supervision (\textit{e.g.}, entity name annotation) of IE stage, and the latter extracts information from only serialized plain text, not using rich visual information.

\begin{figure}
	\centering
	\includegraphics[width=0.5\textwidth]{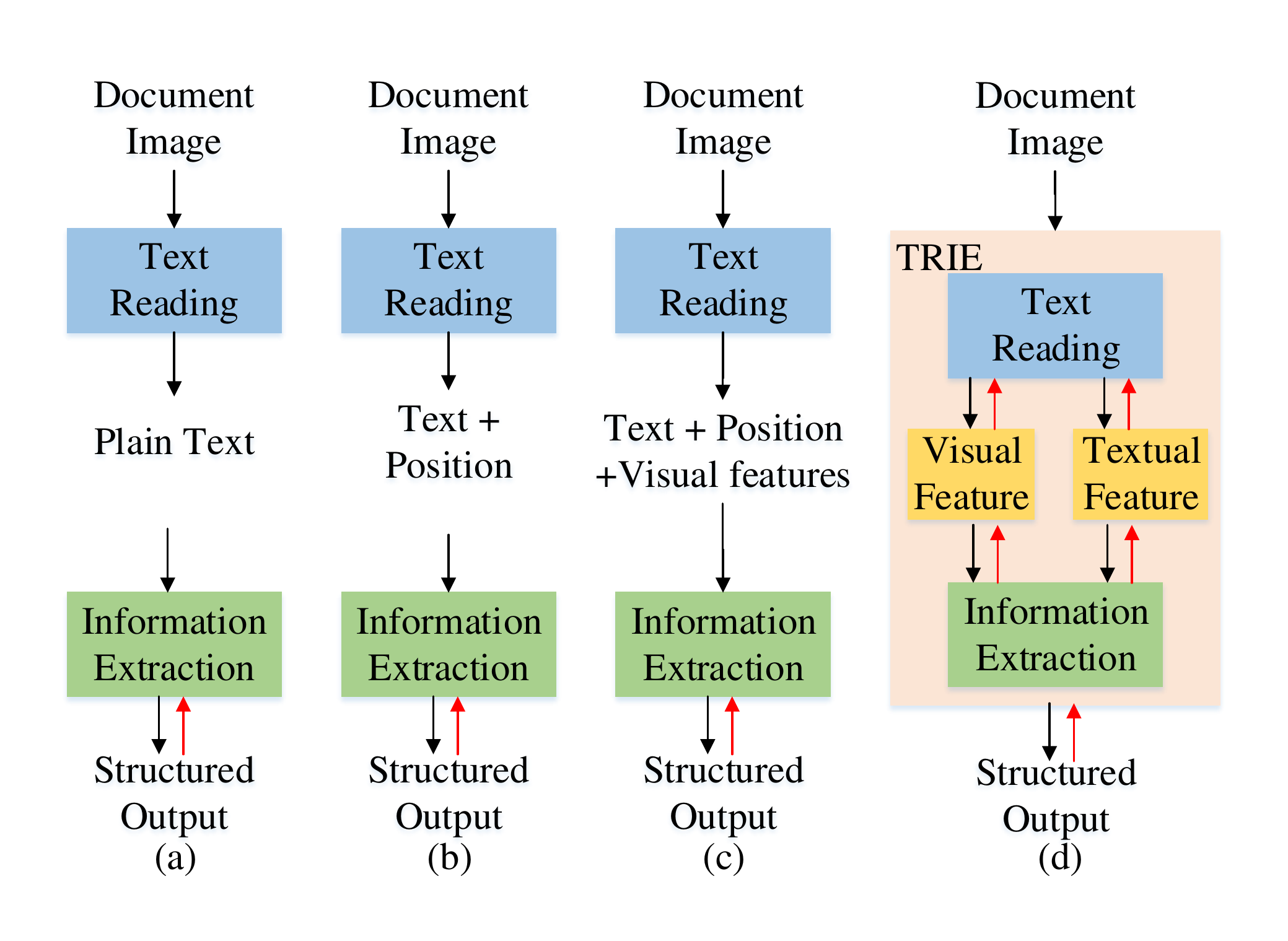}
	\caption{Comparison of VRD understanding architectures: (a) IE with plain text, (b) IE with text and position, (c) IE with position and visual features in addition to plain text, (d) our proposed TRIE. The black and red arrows mean the forward and backward processing, respectively. Best viewed in color.}
	\label{fig:e2e_vs_pipeline}
\end{figure}

Under the traditional routines, most existing methods~\cite{katti2018chargrid,denk2019bertgrid,zhao2019cutie,palm2017cloudscan,liu2019graph,xu2019layoutlm,sage2019recurrent,palm2019attend} design frameworks in multiple stages of text reading (usually including detection and recognition) and information extraction independently in order.
In earlier explorations, the task is intrinsically downgraded into a traditional OCR procedure and the downstream IE from serialized plain text~\cite{palm2017cloudscan,sage2019recurrent}, which completely discards the visual features and layout information from images, as shown in Fig.~\ref{fig:e2e_vs_pipeline}(a).
Noticing the rich visual information contained in document images, recent works incorporate them into network design.
\cite{katti2018chargrid,denk2019bertgrid,zhao2019cutie,palm2019attend,liu2019graph} work on recognized texts and their positions (see Fig.~\ref{fig:e2e_vs_pipeline}(b)), while  \cite{xu2019layoutlm,PICK2020YU} further includes image embeddings (see Fig.~\ref{fig:e2e_vs_pipeline}(c)).
They all focus on the design of information extraction task in VRD understanding.

In essence, all the above works inevitably have the following three limitations.
(1) VRD understanding requires both visual and textual features, but the visual features they exploited are limited.
(2) \textit{Text reading} and \textit{information extraction} are highly correlated, but the relations between them have rarely been explored.
(3) The stagewise training strategy of \textit{text reading} and \textit{information extraction} brings redundant computation and time cost.
	
	To address those limitations, in this paper, we propose a novel end-to-end \emph{Text Reading and Information Extraction} (TRIE) network.
	The workflow is as shown in Fig.~\ref{fig:e2e_vs_pipeline}(d).
	Instead of just focusing on information extraction, we bridge \textit{text reading} and \textit{information extraction} tasks with shared features, including position features, visual features and textual features.
	Thus, the two tasks can reinforce each other amidst a unified framework.
	Specifically, in the forward pass, multimodal visual and textual features produced by \textit{text reading} are fully fused for \textit{information extraction}, while in the backward pass, the semantics in \textit{information extraction} also contribute to the optimization of \textit{text reading}.
	Since all the modules in the network are differentiable, the whole network can be trained end-to-end.
	To the best of our knowledge, this is the first end-to-end trainable VRD framework, which can handle various types of VRDs, from fixed to variable layouts and from structured to semi-structured text type.

	Major contributions are summarized as follows:
	
	(1) We propose an end-to-end trainable framework for simultaneous text reading and information extraction in VRD understanding. 
	The whole framework can be trained end-to-end from scratch, with no need of stagewise training strategies.
	
	(2) We design a multimodal context block to bridge the OCR and IE modules. To the best of our knowledge, it is the first work to mine the mutual influence of text reading and information extraction. 
	
	(3) We perform extensive evaluations on our framework and show superior performance compared with the state-of-the-art counterparts both in efficiency and accuracy on three real-world benchmarks. Note that those three benchmarks cover diverse types of document images, from fixed to variable layouts, from structured to semi-structured text types.

\section{Related Works}
\label{related_work}

Here, we briefly review the recent advances in text reading and information extraction.

\subsection{Text Reading}
Text reading are formally divided into two sub-tasks: text detection and text recognition.  
In text detection, methods are usually divided into two categories: \textit{anchor-based methods} and \textit{segmentation-based methods} .
\textit{Anchor-based methods}~\cite{he2017single, liao2017textboxes, liu2017deep, shi2017detecting,Rosetta18Borisyuk} predict the existence of texts and regress their location offsets at pre-defined grid points of the input image, while \textit{segmentation-based methods}~\cite{zhou2017east,long2018textsnake,Wang2019Shape} learn the pixel-level classification tasks to separate text regions apart from the background. 
In text recognition, the mainstreaming CRNN framework was indroduced by \cite{CRNN}, using recurrent neural networks (RNNs) \cite{LSTM,chung2014empirical} combined with CNN-based methods for better sequential recognition of text lines. 
Then, the attention mechanism replaced existing CTC decoder \cite{CRNN} and was applied to a stacked RNN on top of the recursive CNN \cite{R2AM}, whose performance surpassed the state-of-the-art among diverse variations.

To sufficiently exploit the complementary between detection and recognition,~\cite{liu2018fots,li2017towards,he2018end,busta2017deep, wang2020all,qiao2020textperceptron,feng2019textdragon,MaskTextspotter18Lyu} were proposed to jointly detect and recognize text instances in an end-to-end manner.
They all achieved impressive results compared to traditional pipepline approaches due to capturing \textit{relations} within detection and recognition sub-tasks through joint training. 
This inspires us to dive into the broader field and for the first time pay attention to the relations between text reading and information extraction.  

\subsection{Information Extraction}
Information extraction has been studied for decades.
Before the advent of learning based models, rule-based methods\cite{riloff1993automatically,huffman1995learning,muslea1999extraction,dengel2002smartfix,schuster2013intellix,esser2012automatic} were proposed.
Pattern matching were widely used in~\cite{riloff1993automatically,huffman1995learning} to extract one or multiple target values.
Afterwards, Intellix\cite{schuster2013intellix} required predefined template with relevant fields annotated and SmartFix\cite{dengel2002smartfix} employed specifically designed configuration rules for each template.
Though rule-based methods work in some cases, they rely heavily on the predefined rules, whose design and maintenance usually require deep expertise and large time cost. Besides, they can not generalize across document templates.

Learning-based networks ~\cite{lample2016neural,ma2019end,yadav2019survey,devlin2018bert,dai2019transformer,yang2019xlnet} were firstly proposed to work on plain text documents.
~\cite{palm2017cloudscan,sage2019recurrent} adopted the idea from natural language processing and used recurrent neural networks to extract entities of interest from VRDs.
However, they discard the layout information during the text serialization, which is crucial for document understanding.
Recently, observing the rich visual information contained in document images, works tended to incorporate more details from VRDs. Some works\cite{katti2018chargrid,denk2019bertgrid,zhao2019cutie,palm2019attend} took the layout into consideration, and worked on the reconstructed character or word segmentation of the document. Position information had also been utilized. \cite{liu2019graph} combined texts and their positions through a Graph Convolutional Network (GCN) and \cite{xu2019layoutlm} further integrated position and image embeddings for pre-training inspired by BERT \cite{devlin2018bert}.
{\cite{MajumderPTWZN20} presented a representation learning approach to extract structured information from templatic documents, which worked in pipeline of candidate generation, scoring and assignment.}
However, these works all focus on the design of information extraction task.  
They miss lots of informative details because multi-modality of inputs are not fully explored.

Two related concurrent works were presented in~\cite{qin2019eaten,carbonell2019treynet}.
\cite{qin2019eaten} proposed an entity-aware attention text extraction network to extract entities from VRDs.
However, it could only process documents of relatively fixed layout and structured text, like train tickets, passports and bussiness cards.
\cite{carbonell2019treynet} localized, recognized and classified each word in the document.
Since it worked in the word granularity, it not only required much more labeling efforts (positions, content and category of each word) but also had difficulties in extracting those entities which were embedded in word texts (\textit{e.g.} extracting `51xxxx@xxx.com' from `153-xxx97$|$51xxxx@xxx.com').
Besides, in its entity recognition branch, it still worked on the serialized word features, which were sorted and packed in the left to right and top to bottom order.
Thus, it can only handle documents of simple layout and non-structured text, like paragraph pages.
Both of the two existing works are strictly limited to documents of relative fixed layout and one type of text (structured or semi-structured).
While our proposed framework has the ability to handle documents of both fixed and variable layouts, structured and semi-structured text types.

\section{Methodology}
\label{methodology}
We first introduce the overall architecture of TRIE in Sec \ref{sec3.1}. In the following three sections, text reading module, multimodal context block and information extraction module are illustrated in details respectively. Optimation functions are as shown in Sec \ref{sec3.5}.
\subsection{Overall Architecture}\label{sec3.1}

\begin{figure*}
	\centering
	\includegraphics[width=\textwidth]{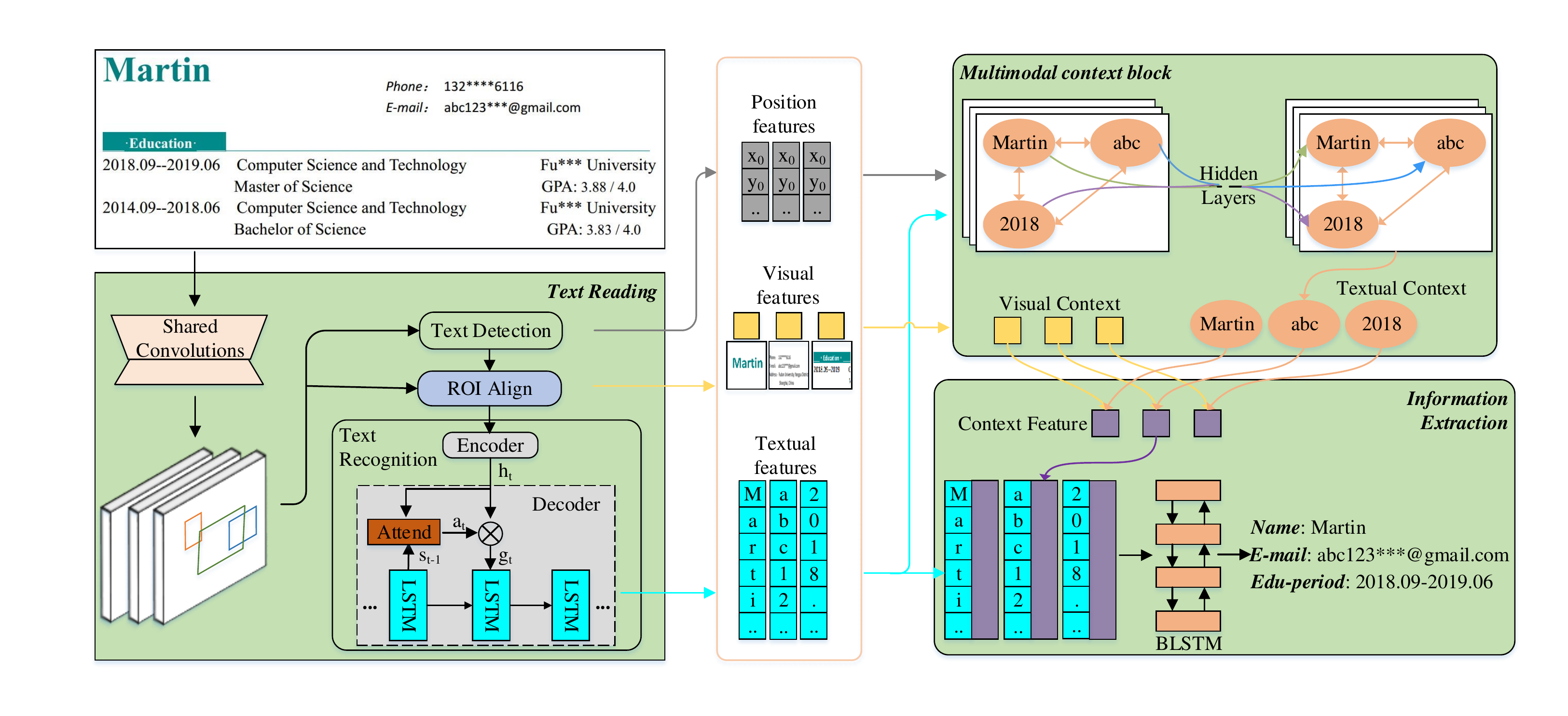}
	\caption{Overall architecture. The network predicts text regions, text content and extract entities of interest in a single forward pass.}
	\label{fig:system_architecture}
\end{figure*}
An overview of the architecture is as shown in Fig.~\ref{fig:system_architecture}.
It mainly consists of three parts: text reading module, multimodal context block and information extraction module. 
{
\textit{Text reading} module is responsible for localizing and recognizing all texts in document images and \textit{information extraction} module is to extract entities of interest from them.
The \textit{multimodal context block} is novelly designed to bridge the text reading and information extraction modules.

Specifically, in text reading,
the network takes the original image as input and outputs text region coordinate positions.
Once the positions obtained, we apply RoIAlign \cite{HeGDG17mask} on the shared convolutional features to get text region features. Then an attention-based sequence recognizor is adopted to get textual features for each text. 
Since context information of a text is necessary to tell it apart from other entities, we design a special \textit{multimodal context block} to provide both visual and textual context.
In the block, the visual context features summarize the local patterns of a text
while textural context features model implicit relations among the whole document image.
These two context features are complementary to each other and fused as final context features.
Finally, entities are extracted through the combined context and textual features.
The whole framework learns to read text and extract entities jointly and can be trained in an end-to-end way from the scratch, which allows bi-directional information flows and brings reinforcing effects among text reading and information extraction modules. 
}

\subsection{Text Reading Module}
Text reading module commonly includes a shared backbone, a text detection branch as well as a sequential recognition branch. 
We adopt ResNet \cite{HeZRS16deep} and Feature Pyramid Network (FPN) \cite{LinDGHHB17feature} as our backbone to extract shared convolutional features.
For an input image $x$, we denote $\mathcal{I} \in \mathbb{R}^{h \times w \times d}$ as the shared feature maps, where $h$, $w$ and $d$ are the height, width and channel number of $\mathcal{I}$.
The text detection branch takes $\mathcal{I}$ as input and predicts the locations of all possible text regions.
\begin{equation}\label{equa1}
\mathcal{B}=\textit{Detector}(\mathcal{I})
\end{equation}
where the $\textit{Detector}$ can be any anchor-based~\cite{he2017single,liao2017textboxes,liu2017deep,shi2017detecting} or segmentation-based ~\cite{zhou2017east,long2018textsnake,Wang2019Shape} text detection methods.
Here, $\mathcal{B}=(b_1, b_2,\dots, b_m)$ is a set of $m$ text bounding boxes and $b_i=(x_0, y_0, $
$x_1, y_1)$ denotes the top-left and bottom-right positions of the $i$-th text.
Given text positions $\mathcal{B}$, RoIAlign \cite{HeGDG17mask} is applied on the shared convolutional features $\mathcal{I}$ to get their text region features, denoted as $\mathcal{C}=(c_1, c_2,\dots, c_m)$, where $c_i \in \mathbb{R}^{h^\prime \times w^\prime \times d}$, $h^\prime$ and $w^\prime$ are the spatial dimensions, and $d$ is the vector dimension the same as in $\mathcal{I}$.

Afterwards, the text recognition branch predicts a character sequence from each text region features $c_i$.
Firstly, $c_i$ is fed into an encoder to extract a higher-level feature sequence $\mathcal{F} \in \mathbb{R}^{l \times d_r}$, where $l$ is the length of the sequence and $d_r$ is vector dimension. 
Then, attention-based decoder is adopted to recurrently generate the sequence of characters $y=(y_1, y_2,\dots, y_T)$, where $T$ is the length of label sequence.
Specifically, at step $t$, the attention weights $\alpha_t$ and glimpse vector $g_t$ are computed as follows,
\begin{equation}
g_t=\sum_{k=1}^{l}\alpha_{t,k}\mathcal{F}_k
\end{equation}
\begin{equation}
\alpha_{t,k}=exp(e_{t,k})/\sum_{j=1}^{l}exp(e_{t,j})
\end{equation}
where,
\begin{equation}
e_{t,j}=w^\mathsf{T}tanh(Ws_{t-1}+M\mathcal{F}_j+b)
\end{equation}
and $w$, $W$, $M$ and $b$ are trainable weights.

The state $s_{t-1}$ is updated via,
\begin{equation}
s_t=RNN(s_{t-1}, g_t, y_{t-1})
\end{equation}
where $y_{t-1}$ is the $(t-1)$-th ground-truth label in training, while in testing, it is the label predicted in the previous step.
In our experiment, LSTM \cite{LSTM} is adopted as RNN unit.
Finally, the probability distribution over the vocabulary label space is estimated by,
\begin{equation}
p_{t}^{rcg}=softmax(W^{rcg}s_t + b^{rcg})
\end{equation}
where $W^{rcg}$ and $b^{rcg}$ are learnable weights.
We denote that $z_i=(s_1, s_2, \dots, s_T)$ as the textual features of $i$-th text, as they contain semantic features for each character in it.

In a nutshell, the text reading module outputs visual features $\mathcal{C}=(c_1, c_2,\dots, c_m)$ and textual features $\mathcal{Z}=(z_1, z_2, \dots, z_m)$ of $m$ texts in the image in addition to their positions $\mathcal{B}=(b_1, b_2,\dots, b_m)$.

\subsection{Multimodal Context Block}
The context of a text provides necessary information to tell it apart from other entities, which is crucial for information extraction.
Unlike the most existing works which rely only on textual features and/or position features, we design a multimodal context block to consider position features, visual features and textual features all together.
This block provides both visual context and textual context of a text, which are complementary to each other and further fused in the information extraction module.
 
\subsubsection{Visual Context.}
As mentioned, visual details such as the obvious color, font, layout and other informative features are equally important as textual details (text content) for document understanding.
A natural way of capturing the local visual context of a text is resort to the convolutional neural network.
Different from \cite{xu2019layoutlm} which extracts these features from scratch, we directly reuse $\mathcal{C}=(c_1, c_2, \dots, c_m)$ produced by the text reading module.
Thanks to the deep backbone and lateral connections introduced by FPN, each  $c_i$ summarizes the rich local visual patterns of the $i$-th text.

\subsubsection{Textural Context.}
Unlike visual context which focuses on local visual patterns, textual context models the fine-grained long distance dependencies and relationships between texts, providing complementary context information.
Inspired by \cite{devlin2018bert,VisualBERTLi,Lu2019ViLBERT}, we apply the self-attention mechanism to extract textual context features, supporting variable number of texts.

\textbf{Self-attention Recap.}
The input of popular scaled dot-product attention consists of queries $Q$ and keys $K$ of dimension $d_k$, and values $V$ of dimension $d_v$.
The output is obtained by weighted summation over all values $V$ and the attention weights are learned from $Q$ and $K$. Please refer to \cite{devlin2018bert} for details.

\textbf{Textual Context Modeling.}
To retain the document layout and content as much as possible, we make the use of positions $\mathcal{B}=(b_1, b_2, \dots, b_m)$ and textual features $\mathcal{Z}=(z_1, z_2, \dots, z_m)$ at the same time.
We first introduce \textit{position embeddings} to preserve  layout information,
Then, a $ConvNet$ of multiple kernels similar to~\cite{zhang2015character} is used to aggregate semantic character features in $z_i$ and outputs $\widehat{z_i}$,
\begin{equation}
\widehat{z_i}=ConvNet(z_i).
\end{equation}
Finally, the $i$-th text's embedding is fused of $\widehat{z_i}$ and position embedding, followed by the $LayerNorm$ normalization, defined as
\begin{equation}
embedding_i=LayerNorm(Linear(\widehat{z_i}) + PE(b_i))
\end{equation}
where, $Linear$ is a fully-connected layer, projecting $\widehat{z_i}$ to the same dimension with $PE(b_i)$. $b_i$ is the bounding box coordinates of $i$-th text (computed by Equa. \ref{equa1}) and $PE$ is the position embedding layer.
Afterwards, we pack all the texts' embedding vector together, which serves as the $K$, $Q$ and $V$ in the scaled dot-product attention. 

The textual context features $\widetilde{\mathcal{C}}=(\widetilde{c_1}, \widetilde{c_2}, \dots, \widetilde{c_m})$ is obtained by,
\begin{equation}
\begin{split}
\widetilde{\mathcal{C}}&=Attention(Q,K,V) \\
&=softmax(\frac{QK^\mathsf{T}}{\sqrt{d_{info}}})V
\end{split}
\end{equation}
where $d_{info}$ is the dimension of text embeddings, and $\sqrt{d_{info}}$ is the scaling factor.
To further improve the representation capacity of the attended feature, multi-head attention is introduced. Each head corresponds to an independent scaled dot-product attention function and the text context features $\widetilde{\mathcal{C}}$ is given by:
\begin{equation}
\begin{split}
\widetilde{\mathcal{C}}&=MultiHead(Q,K,V)\\
&=[head_1, head_2, ..., head_n]W^{info}
\end{split}
\end{equation}
\begin{equation}
head_j=Attention(QW_j^Q, KW_j^K, VW_j^V)
\end{equation}
where $W^Q_j$, $W^K_j$ and $W^V_j$ $\in \mathbb{R}^{(d_{info}\times d_n)}$ are the learned projections matrics for the $j$-th head, $n$ is the number of heads, and $W^{info}\in \mathbb{R}^{(d_{info} \times d_{info})}$. To prevent the multi-head attention model from becoming too large, we usually have $d_n = \frac{d_{info}}{n}$.

\subsection{Information Extraction Module}
Both the context and textual features matter in entity extraction.
The context features (including both visual context features $\mathcal{C}$ and textual context features $\widetilde{\mathcal{C}}$) provide necessary information to tell entities apart while the textual features $\mathcal{Z}$ enable entity extraction in the character granularity, as they contain semantic features for each character in the text.
So we first perform multimodal fusion of visual context features $\mathcal{C}$ and textual context features $\widetilde{\mathcal{C}}$, which are further combined with textual features $\mathcal{Z}$ to extract entities.

\subsubsection{Adaptive Multi-modality Context Fusion}
Given the visual context features $\mathcal{C} \in \mathbb{R}^{m \times h^\prime \times w^\prime \times d}$ and textual context features $\widetilde{\mathcal{C}}  \in \mathbb{R}^{m \times d_{info}}$, we fuse them adaptively to get the final context of each text.
Specifically, for the $i$-th text, $c_i \in \mathbb{R}^{ h^\prime \times w^\prime \times d}$ is spatially averaged to output the visual context vector.
The final context vector $\overline{c_i}$ is a weighted sum of visual context vector and textual context vector. That is, $\widetilde{c_i}$
\begin{equation}
\overline{c_i}=\alpha \cdot Linear(\frac{1}{h^\prime w^\prime}\sum_{j}^{h^\prime w^\prime}c_{ij}) + \beta \cdot Linear(\widetilde{c_i}),
\end{equation}
where $\alpha$ and $\beta$ are automatically learnable weights. $Linear$ is a fully-connected layer, projecting the visual context vector and text context vector to the same dimension.

\subsubsection{Entity Extraction}
We further combine the context vector $\overline{c_i}$ and textual features $z_i$ by concatenating $\overline{c_i}$ to each $s_j$ in $z_i$,
\begin{equation}
U_i=(u_{1,1}, u_{i,2}, \dots, u_{i,T}),\ where\ u_{j}=s_{i,j}||\overline{c_i}
\end{equation}
Then, a Bidirectional-LSTM is applied to further model the dependencies within the characters,
\begin{equation}
H_{i}=(h_{i,1}, h_{i,2}, \dots, h_{i,T}) = BiLSTM(U_i),
\end{equation}
which is followed by a fully connected network, projecting the output to the dimension of IOB \cite{SangV99representing} label space.
\begin{equation}
p_{i,j}^{info} = softmax(Linear(h_{i,j}))
\end{equation} 

\subsection{Optimization}\label{sec3.5}

The proposed network can be trained in an end-to-end manner and the losses are generated from three parts,
\begin{equation}
\label{losses}
\mathcal{L}=\mathcal{L}_{det} + \lambda_{recog}\mathcal{L}_{recog} + \lambda_{info}\mathcal{L}_{info}
\end{equation}
where hyper-parameters $\lambda_{recog}$ and $\lambda_{info}$ control the trade-off between losses.

$\mathcal{L}_{det}$ is the loss of text detection branch, which consists of a classification loss and a regression loss, as defined in \cite{RenHG017}. 
The recognition loss and information extraction loss are formulated as,
\begin{equation}
\mathcal{L}_{rcg}=-\frac{1}{T}\sum_{i=1}^{m}\sum_{t=1}^{T}log\ p^{rcg}_{i,t}(y_{i,t}^{rcg})
\end{equation}
\begin{equation}
\mathcal{L}_{info}=-\frac{1}{T}\sum_{i=1}^{m}\sum_{t=1}^{T}log\ p^{info}_{i,t}(y_{i,t}^{info})
\end{equation}
where $y_{i,t}^{rcg}$ is the ground-truth label of $t$-th character in $i$-th text from recognition branch and $y_{i,t}^{info}$ is its corresponding label of information extraction.

Note that since \textit{text reading} and \textit{information extraction} modules are bridged with shared visual and textual features, they can reinforce each other.
Specifically, the visual and textual features of text reading are fully fused and essential for information extraction, while the semantic feedback of information extraction also contributes to the optimization of the shared convolutions and text reading module.

\section{Experiments}
\label{experiment}

In this section, we perform experiments to verify the effectiveness of the proposed method.

\subsection{Datasets}
\begin{table}
	\begin{center}
		\caption{Statistics of benchmark datasets used in this paper.}
		\label{table:datasets}
		\scalebox{0.9}{\begin{tabular}{cccccc}
			\toprule
			Dataset    &  Training    &  Testing  & Entities & Layout & \makecell[c]{Text Type}\\
			\midrule
			\makecell[c]{Taxi Invoices} & 4000 & 1000  &  9 & Fixed & Struct\\
			SROIE & 626 & 347 &  4 & Variable & Struct\\
			Resumes & 1978 & 497 &  6 & Variable & \makecell[c]{Semi-struct}\\
			\bottomrule
		\end{tabular}}
	\end{center}
\end{table}
 
We validate our model on three real-world datasets. One is the public SROIE \cite{HuangCHBKLJ19competition} benchmark, and the other two are self-built datasets, Taxi Invoices and Resumes, respectively.
Note that, the three benchmarks differ largely in layout and text type, from fixed to variable layout and from structured to semi-structured text. Statistics of the datasets are listed in Table~\ref{table:datasets} and some examples are shown in Fig.~\ref{fig:dataset}.
\begin{itemize}
\item \textit{Taxi Invoices} consists of 5000 images and has 9 entities to extract (Invoice Code, Invoice Number, Date, Pick-up time, Drop-off time, Price, Distance, Waiting, Amount).
The invoices are in Chinese and can be grouped into roughly 13 templates.
So it is kind of document of fixed layout and structured text type.
\item \textit{SROIE} \cite{HuangCHBKLJ19competition} is a public dateset for receipt information extraction in ICDAR 2019 Chanllenge.
It contains 626 receipts for training and 347 receipts for testing.
Each receipt is labeled with four types of entities, which are Company, Date, Address and Total.
It has variable layouts and structured text.
\item \textit{Resumes} is a dataset of 2475 Chinese scanned resumes, which has 6 entities to extract (Name, Phone Number, Email Address, Education period, Universities and Majors).
As an owner can design his own resume template, this dataset has variable layouts and semi-structured text.
\end{itemize}

\subsection{Implementation Details} 
\subsubsection{Network Details:}\label{sec4.2.1}

The backbone of our model is ResNet-50 \cite{HeZRS16deep}, followed by the FPN \cite{LinDGHHB17feature} to further enhance features.
The text detection branch in \textit{text reading module} adopts the Faster R-CNN \cite{RenHG017} network and outputs the predicted bounding boxes of possible texts for later
sequential recognition.
For each text region, its features of shape $32\times256$ are extracted from the shared convolutional features by RoIAlign \cite{HeGDG17mask} and
decoded by LSTM-based attention \cite{cheng2017focusing}, where the number of hidden units is set to 256.
In the \textit{information extraction module}, we use convolutions of three kernel size $[3, 5, 7]$ followed by max pooling to extract representations of text.
The dimension of text's embedding vector is 256.
The textual context modeling module consists of multiple layers of multi-head scaled dot-product attention modules.
The number of hidden units of BiLSTM used in entity extraction is set to 128.
Hyper-parameters $\lambda_{recog}$ and $\lambda_{info}$ in Equa. \ref{losses} are all empirically set to 1 in our experiments.

\subsubsection{Training:}
Our model and all the reimplemented counterparts are implemented under the PyTorch framework \cite{paszke2019pytorch}.
For our model, with the ADADELTA~\cite{zeiler2012adadelta} optimization method, we set the learning rate to 1.0 at the beginning and decreased it to a tenth every 40 epoches.
The batch size is set to 2 images per GPU and we train our model for 150 epoches.
For compared counterparts, we train the separate text reading and information extraction tasks independently until they fully converge.
All the experiments are carried out on a workstation with two NVIDIA Tesla V100 GPUs.

\subsection{State-of-the-Art Comparisons}
We compare our proposed method with pipeline of SOTA text reading and information extraction.
\subsubsection{Network Settings of SOTA counterparts and Evaluation Metric.} 
Most existing works only focus on information extraction task for document understanding.
For fair comparisons, we train the \textit{text reading} network which is identical to our model individually and take the results as the input of information extraction counterparts.
We re-implement three top information extraction methods \cite{katti2018chargrid, ma2019end,liu2019graph}. Their network settings are as follows: 
\begin{enumerate}
	\item \textit{\textbf{Chargrid(TR):} Pipeline of text reading and Chargrid}. For \emph{Chargrid}\cite{katti2018chargrid}, the input character embedding is set to $128$ and the rest of network is identical to the paper.
	\item \textit{\textbf{NER(TR):} Pipeline of text reading and LSTM-CRF}. The input embedding of \emph{LSTM-CRF}\cite{ma2019end} is set to $256$, followed by a BiLSTM with $128$ hidden units and CRF.
	\item \textit{\textbf{GCN(TR):} Pipeline of text Reading and GCN}. In \emph{GCN}, we build the network as the paper \cite{liu2019graph} specified and set the input embedding to $256$, which performs slightly better than $128$.
\end{enumerate} 
In consistence with \cite{liu2019graph} and the SROIE Challenge \cite{HuangCHBKLJ19competition}, we use F1-score to evaluate the performance of all experiments.

\subsubsection{Results.}
\begin{figure}
	\centering
	\includegraphics[width=0.30\textwidth]{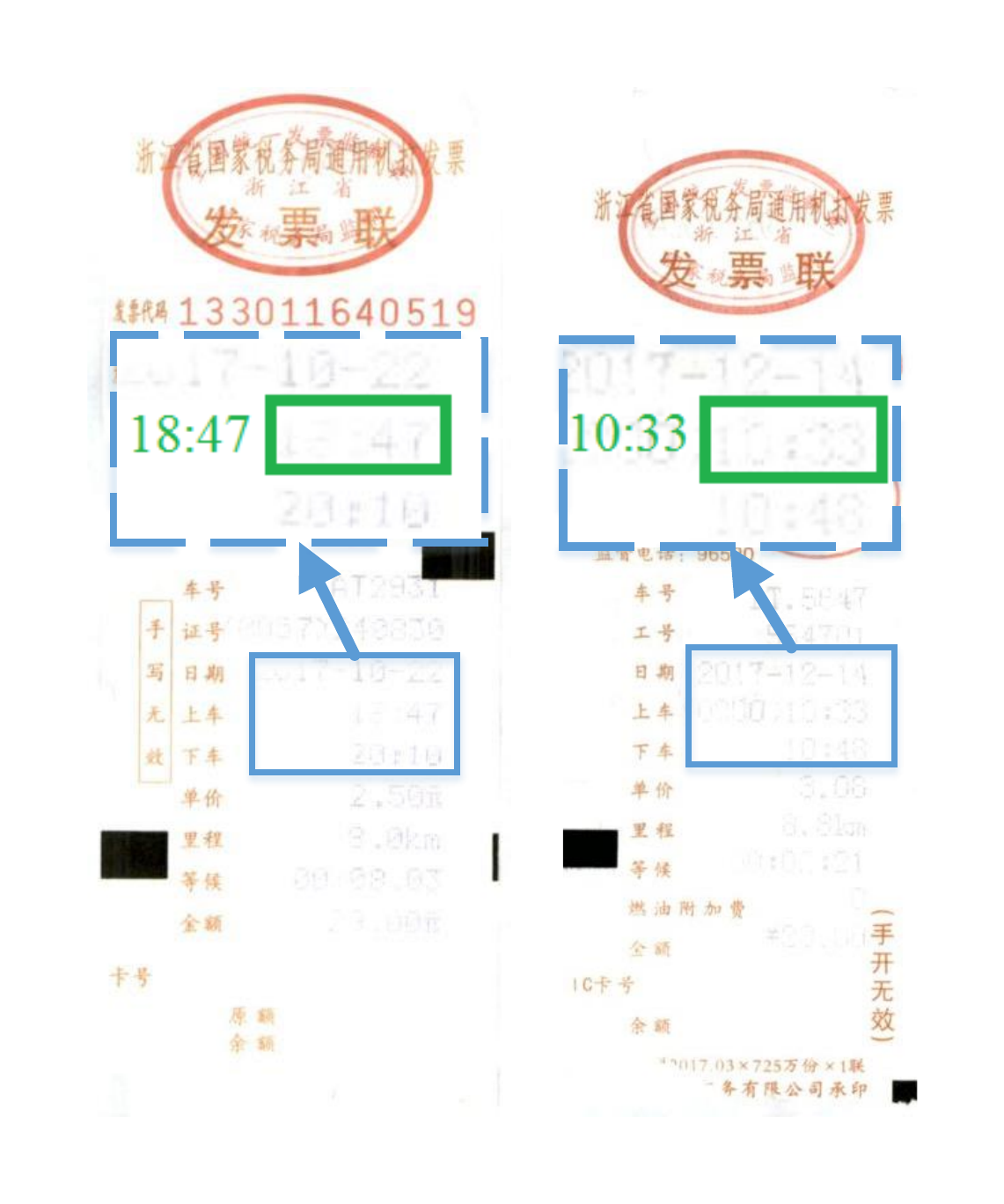}
	\caption{The Pick-up time which is too blur to read is annotated as NULL, but our model correctly read and extract it out, which results in lower accuracy.}
	\label{fig:errors}
\end{figure}

\textbf{Evaluation on Taxi Invoices:} 
In this dataset, the noise of low-quality and taint may lead to failures of detection and recognition of entities.
Besides, the contents may be misplaced, \textit{e.g.} the content of Pick-up time may appear after the `Date'.
Table~\ref{table:performance-taxi} shows the results.
We see that our model outperforms counterparts by significant margins except for the Pick-up time (illustrated in the tail of the paragraph).
Concretely,
\textit{NER(TR)} discards the layout information and serializes all texts into one-dimensional text sequences, reporting inferior performance than other methods.  
Benefiting from the layout information,  
\textit{Chargrid(TR)} and \textit{GCN(TR)} work much better.
However, \textit{Chargrid(TR)} conducts pixel segmentation task and is prone to omit characters or include extra characters.
For \textit{GCN(TR)}, it only exploits the positions of text segments.
Obviously, our \textit{TRIE} has the ability to boost performances by using more useful visual features in VRDs.
In addition, we attribute the only slight lower score of Pick-up entity compared with \textit{GCN(TR)} to the annotations.
For example in Fig.~\ref{fig:errors}, when an entity such as the Pick-up time `18:47' is too blurred to read, it is tagged as NULL. However, our model can still correctly read and extract this entity out, which leads to lower statistics.
\begin{table}
	\begin{center}
		\caption{Performance comparisons (F1-score) on Taxi Invoices.}
		\label{table:performance-taxi}
		\scalebox{0.9}{\begin{tabular}{lcccc}
			\toprule
			Entities    &  Chargrid(TR)     &  NER(TR)   & GCN(TR) & Our Model\\
			\midrule
			Code & 89.4 & 94.5 & 97.0 & \textbf{98.2} \\
			Number & 85.3 & 92.4 & 93.7 & \textbf{95.4} \\
			Date & 89.8 & 82.5 & 93.0 & \textbf{94.9} \\
			Pick-up time & 82.9 & 60.0 & \textbf{86.3} & 84.6 \\
			Drop-off time & 87.4 & 81.1 & 91.0 & \textbf{93.6} \\
			Price & 93.0 & 94.5 & 93.6 & \textbf{94.9} \\
			Distance & 92.7 & 93.6 & 91.4 & \textbf{94.4} \\
			Waiting & 89.2 & 85.4 & 91.0 & \textbf{92.4} \\
			Amount & 80.2 & 86.3 & 88.7 & \textbf{90.9} \\
			\midrule
			Avg & 87.77 & 85.59 & 91.74 & \textbf{93.26} \\
			\bottomrule
		\end{tabular}}
	\end{center}
\end{table}

\textbf{Evaluation on SROIE:}
We perform two sets of experiments and the results are as shown in Table~\ref{table:performance-sroie}.

\emph{Setting 1}: We train text reading module all by ourselves and report comparisons.
Notice that, we do \emph{not} employ tricks of data synthesis and model ensemble in the training of text reading. 
Since entities of `Company' and `Total' often have distinguishing visual features (\emph{e.g.,} bold type or large font), as shown in~Fig. \ref{fig:dataset}(b),
benefiting from fusion of visual and textual features, our model outperforms three counterparts by a large margin.

\emph{Setting 2}: For fair comparisons on the leadboard, we use the groundtruth of text bounding-box and transcript provided officially.
\emph{Character-Word LSTM} is similar to NER \cite{lample2016neural}, which applies LSTM on character and word level sequentially. 
LayoutLM~\cite{xu2019layoutlm} makes use of large pre-training data and fine-tunes on SROIE.
{Similar to LayoutLM, PICK~\cite{PICK2020YU} extracts rich semantic representation containing the textual and visual features as well as global layout.}
Compared with these methods, our model shows competitive performance.
\begin{table}
	\begin{center}
		\caption{Performance comparisons (F1-score) on SROIE.}
		\label{table:performance-sroie}
		\scalebox{0.95}{\begin{tabular}{l|l|c}
			\hline
			Setting                         & Model              & \multicolumn{1}{l}{F1-Score} \\ \hline
			\multirow{4}{*}{\makecell[l]{{Setting 1}: \\Prediction of bboxes \\and transcript of texts}} & Chargrid(TR) &  78.24\\
			& NER(TR)  & 69.09 \\
			& GCN(TR)  &  76.51\\
			& Our model          &  \textbf{82.06} \\ \hline
			\multirow{4}{*}{\makecell[l]{{Setting 2}: \\Groundtruth of bboxes \\and transcript of texts}}   & Character-Word LSTM \cite{lample2016neural} &  90.85\\
			& LayoutLM\cite{xu2019layoutlm} & 95.24 \\
			& PICK\cite{PICK2020YU} & 96.12 \\
			& Our model          &  \textbf{96.18} \\ \hline
		\end{tabular}}
	\end{center}
\end{table}

\textbf{Evaluation on Resumes:}
On the variable layout and semi-structured dataset, our method still shows impressive results. Table~\ref{table:performance-resume} shows the results.
We find that \textit{GCN(TR)} has difficulty in adapting to such flexible layout. 
\textit{Chargrid(TR)} obtains impressive performances on isolated entities such as Name, Phone and Education period.
Since University and Major entities are often blended with other texts, \textit{Chargrid(TR)} may be failed to extract these entities.
As expected, \textit{NER(TR)} performs better on this dataset, thanks to the inherent serializable property.
While it is inferior to \textit{Chargrid(TR)} on isolated entities due to the missing of layout information.
Our model inherits the advantages of both \textit{Chargrid(TR)} and \textit{NER(TR)}, providing context features for identifying entities and performing entity extraction in the character-level.
In short, our model gets comprehensive gain. 
\begin{table}
	\begin{center}
		\caption{Performance comparisons (F1-score) on Resumes.}
		\label{table:performance-resume}
		\scalebox{0.9}{\begin{tabular}{lcccc}
			\toprule
			Entities    & Chargrid(TR)     &  NER(TR)   & GCN(TR) & Our Model\\
			\midrule
			Name     &  43.4     &  42.7     &  42.9     &  \textbf{45.7}   \\
			Phone     & 87.0  &  86.6     &  83.3     & \textbf{88.0}    \\
			E-mail     &  70.9  &  69.6     &  68.0     & \textbf{74.9}    \\
			Edu-period     &  77.1 &  68.7     & 62.2      & \textbf{81.4}    \\
			University     &  74.7 &  86.0     &  82.3     &  \textbf{87.4}   \\
			Major     &  72.1 &  80.4     &  78.7     & \textbf{80.8}    \\
			\midrule
			Avg & 70.87 & 72.33 & 69.57 & \textbf{76.3} \\
			\midrule
			Speed(fps) & 1.13 & 1.69 & 1.62 & \textbf{1.76} \\
			\bottomrule
		\end{tabular}}
	\end{center}
\end{table}

\subsection{Discussion} 
\label{Discussion}
In this section, we offer more detailed discussion and deeper understanding on different part of the proposed \textit{TRIE} framework.
\subsubsection{Effect of multi-modality features on information extraction:}
\label{forward_effect}
\begin{figure}
	\centering
	\scalebox{0.95}{\includegraphics[width=0.5\textwidth]{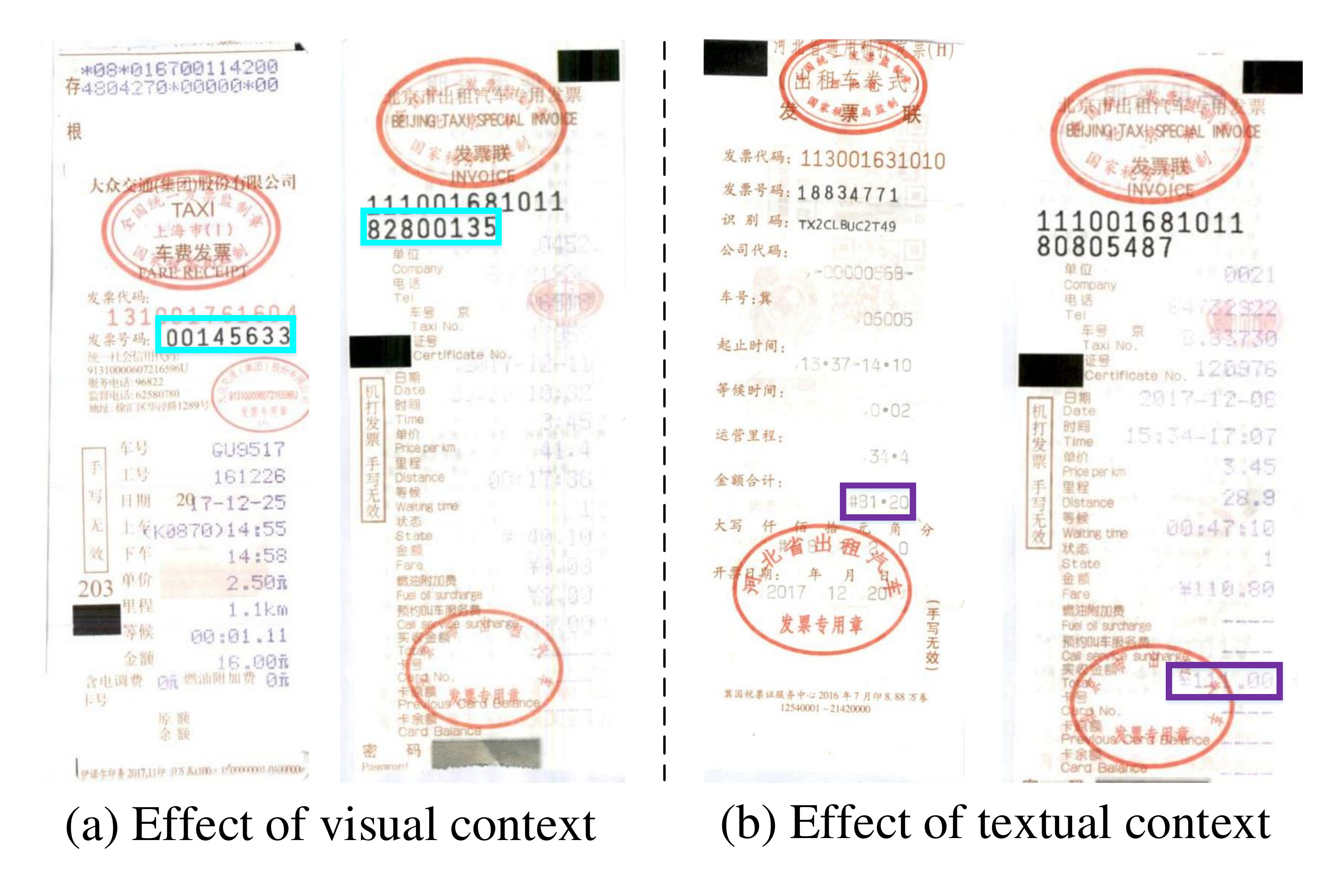}}
	\caption{Illustration of effects of visual and textual context (highlighted with box). Taxi Code in (a) has distinguishing visual features, such as bold type and large font, which can boost information extraction performance. Invoices in (b) have unusual layouts where the visual context often fails. The textual context models dependencies and relationships between texts.}
	\label{fig:context_contribution}
\end{figure}

To further examine the contributions of visual and textual context features to information extraction, we perform the following ablation study on the Taxi Invoices and the result is shown in Table~\ref{table:performance-contribution}.
\emph{Text feat only} means that entities are extracted using features from text reading module only.
Since the layout information is completely lost, this method presents worst performance.
Introducing either the \emph{visual context features} or \emph{textual context features} brings significant performance gains.
Notice that, introducing \emph{visual context features} outperforms \emph{textual context features} slightly, revealing the effects of visual context features.
Further fusion of the above multi-modality features gives the best performance, which verifies the effect of multi-modality features.
{Visual examples in Fig.~\ref{fig:context_contribution} illustrates the effects of each piece of context.}
\begin{table}
	\begin{center}
		\caption{Effects of multimodal features on information extraction.}
		\label{table:performance-contribution}
		\begin{tabular}{lcccc}
			\toprule
			Text feat only & $\surd$ & $\surd$& $\surd$& $\surd$\\
			Textual Context feat & & $\surd$ & & $\surd$\\
			Visual Context feat & & & $\surd$ & $\surd$\\
			\midrule
			Accuracy & 74.33 & 92.30 & 92.70 & \textbf{93.26}\\
			\bottomrule
		\end{tabular}
	\end{center}
\end{table}

\subsubsection{Effect of the end-to-end framework on text reading:}  
\begin{figure}
	\centering
	\scalebox{0.95}{\includegraphics[width=0.5\textwidth]{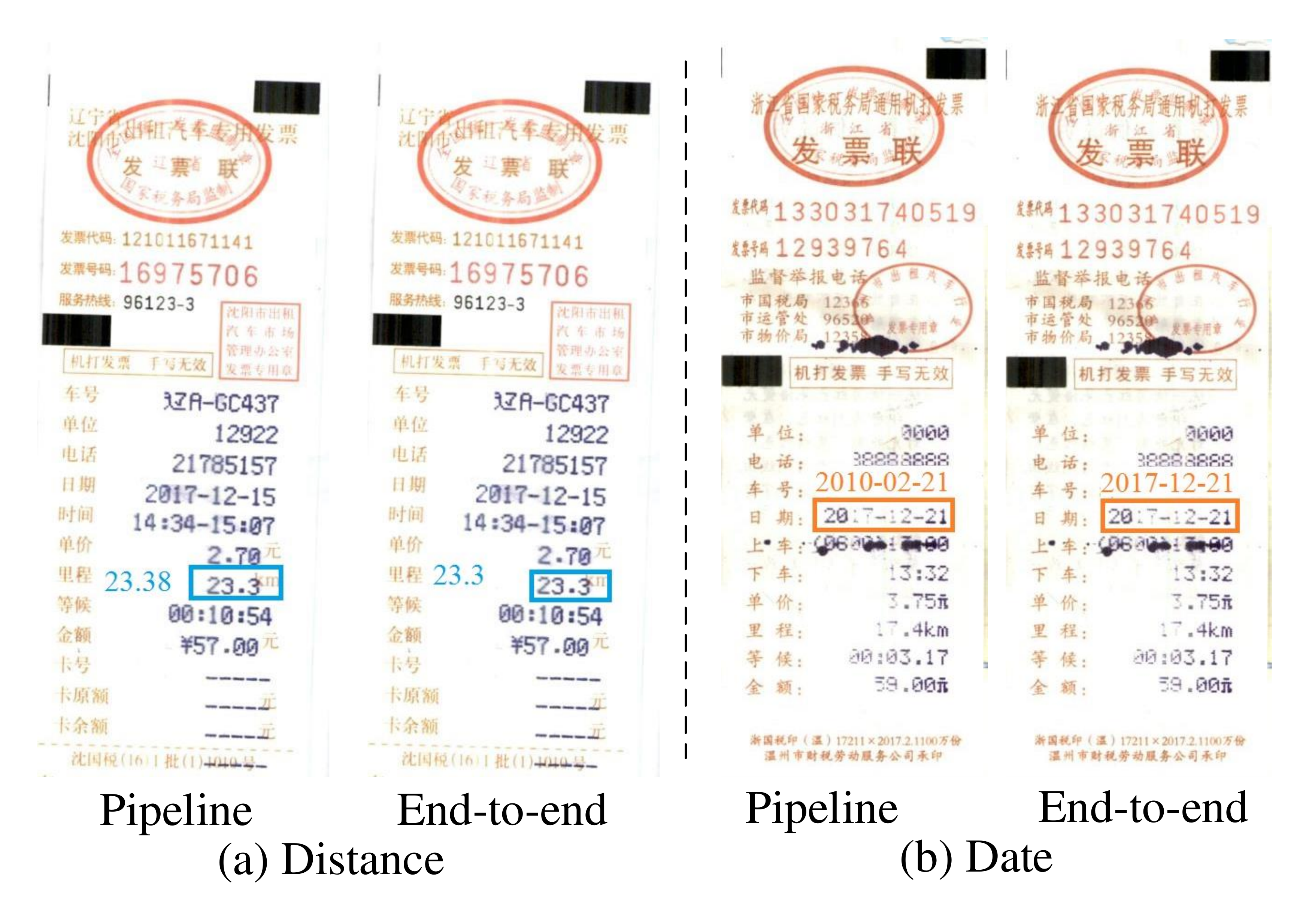}}
	\caption{Visualization of text reading results of pipeline training and our end-to-end training. Best viewed in color.}
	\label{fig:res_vs_pipeline_e2e}
\end{figure}

To verify the effect of end-to-end framework on text reading, we evaluate GCN \cite{liu2019graph} model on two sets of text reading results, and the comparison results are shown in Table~\ref{table:performance-e2e-vs-pipeline}.
\textit{TR only} means training text reading module individually, and \textit{End-to-End (TRIE)} is our proposed model.
GCN on results of \textit{End-to-End (TRIE)} shows $0.9$ F1-score gain over that on \textit{TR only}, demonstrating the effectiveness of end-to-end training. 
Some visualization examples are shown in Fig.~\ref{fig:res_vs_pipeline_e2e}.
The Distance `23.3km' and Date `2017-12-21' are incorrectly recognized as `23.38m' and `2010-02-21' in pipeline training, leading to wrong information extraction results. While we can obtain their correct results when evaluating in an end-to-end training way.
These quantitative and qualitative results uncover the effect of end-to-end framework on text reading.
Considering effects of multimodal features (See Section \ref{forward_effect}), we can conclude that joint modeling of text reading and information extraction can benefit both tasks. 
\begin{table}
	\begin{center}
		\caption{Effect of end-to-end framework on text reading.}
		\label{table:performance-e2e-vs-pipeline}
		\begin{tabular}{cc|cc}
			\toprule
			\multicolumn{2}{c|}{Text Reading Results} &\multirow{2}{*}{IE Model}& \multirow{2}{*}{Accuracy}\\
			TR only & End-to-End (TRIE) & &  \\
			\midrule
			$\surd$ &&GCN \cite{liu2019graph}& 91.70 \\
			&$\surd$ &GCN \cite{liu2019graph}& 92.60 \\
		    &$\surd$ &TRIE& 93.26\\
			\bottomrule
		\end{tabular}
	\end{center}
\end{table}

\subsubsection{Speed:}
We evaluate the running time of our model and its counterparts in \textit{fps} (frames per second).
Results are shown in Table~\ref{table:performance-resume}.
Since \textit{text reading} needs to detect and recognize all the texts in images, it takes up most of the time and runs at $1.83$ fps.
Compared to \textit{text reading}, the \textit{information extraction} is much more lightweight.
That is, \textit{Chargrid}, \textit{NER} and \textit{GCN} run at $2.97$, $22.40$ and $14.58$ fps, respectively.
In total, their full pipeline \textit{Chargrid(TR)}, \textit{NER(TR)} and \textit{GCN(TR)} report speed of $1.13$, $1.69$ and $1.62$ respectively.
Benefiting from feature sharing between \textit{text reading} and \textit{information extraction}, \textit{TRIE} achieves the highest speed of $1.76$ fps.
Note that, all methods are tested on Resumes. 
\textit{Text reading } uses $2666\times1600$ size image as input and $32\times512$ image patches for recognition. 
$Chargrid$ resolution is set to $1344\times950$ to reduce computations. 

\subsubsection{Ablation on layers and heads in Textural Context Block:} 
To analyze the effects of different numbers of layers and heads in the \emph{Textural Context Block}, we perform the following ablation studies, and the results are shown in Table~\ref{table:performance-layers-heads}.
Taxi Invoices is relatively simple and has fixed layout, thus model with 1 or 2 layers and small number of heads gives the best result.
As the layers and heads grow deeper, the model is prone to overfit.
The layout of Resumes is much more variable than Taxi Invoices, thus it requires deeper layers and heads.
In practice, one can adjust these settings according to the complexity of a specific task. 
\begin{table}[]
	\begin{center}
		\caption{Effects of Layers and Heads in Textural Context Block.}
		\label{table:performance-layers-heads}
		\begin{tabular}{l|c|cccc}
			\hline
			\multirow{2}{*}{Datasets}      & \multirow{2}{*}{Layers} & \multicolumn{4}{c}{Heads} \\ \cline{3-6}
			&   &2                & 4      & 8      & 16      \\ \hline
			\multirow{3}{*}{\makecell[l]{Taxi \\Invoices}} & 1  & 92.97 & 92.98 & 92.86 & 92.72 \\
			& 2  & 93.00 & 92.98 & \textbf{93.26} &  92.71   \\
			& 3  & 92.55 & 92.81 & 93.06  &  92.83  \\ \hline
			\multirow{3}{*}{Resumes}       & 1 &  75.20 &   75.21     &  75.47     &  74.53       \\
			& 2 &   75.62  &    76.25    &  76.28      &  75.86       \\
			& 3 &   75.55  &  75.74      &  \textbf{76.35}      &  76.35       \\ \hline
		\end{tabular}
	\end{center}
\end{table}

\section{Conclusion}
\label{conclusion}
In this paper, we presented an end-to-end network to bridge the text reading and information extraction for document understanding.
These two tasks can mutually reinforce each other through joint training.
The visual and textual features of text reading can boost the performances of information extraction while the loss of information extraction can also supervise the optimization of text reading.
On a variety of benchmarks, from structured to semi-structured text type and fixed to variable layout, our proposed method significantly outperforms three state-of-the-art methods in both aspects of efficiency and accuracy.

\bibliographystyle{ACM-Reference-Format}
\balance
\bibliography{egbib}

\end{document}